\documentclass[letterpaper]{article} 
\usepackage[preprint]{aaai2027}  
\usepackage[hyphens]{url}  
\usepackage{graphicx} 
\usepackage{natbib}  
\usepackage{caption} 
\usepackage{algorithm}
\usepackage{algorithmic}
\usepackage{newfloat}
\usepackage{listings}
\DeclareCaptionStyle{ruled}{labelfont=normalfont,labelsep=colon,strut=off} 
\floatstyle{ruled}
\newfloat{listing}{tb}{lst}{}
\floatname{listing}{Listing}

\usepackage{booktabs}
\usepackage{amsmath}
\usepackage{amssymb}

\title{PosterText: Towards Unified Visual Text Generation and Editing \\ for E-commerce Poster}
\author{
    Xiaoan Liu\textsuperscript{\rm 1,\rm 2$*$},
    Lichen Ma\textsuperscript{\rm 2,\rm 3$*$},
    Zipeng Guo\textsuperscript{\rm 2$*$},
    Yu He\textsuperscript{\rm 2},
    Xiaoyan Su\textsuperscript{\rm 4},
    Shaojie Guo\textsuperscript{\rm 2},
    Jingling Fu\textsuperscript{\rm 2},\\
    Xiaolong Fu\textsuperscript{\rm 2},
    Hao Yang\textsuperscript{\rm 3},
    Tongxuan Liu\textsuperscript{\rm 2},
    Yu Guo\textsuperscript{\rm 3},
    Fei Wang\textsuperscript{\rm 3},
    Xinyi Liu\textsuperscript{\rm 1},
    Yongjun Zhang\textsuperscript{\rm 1$\ddagger$},\\
    Junshi Huang\textsuperscript{\rm 2$\dagger$}
}
\affiliations{
    \textsuperscript{\rm 1}Wuhan University,
    \textsuperscript{\rm 2}JD.com,
    \textsuperscript{\rm 3}State Key Laboratory of Human-Machine Hybrid Augmented Intelligence,
    Institute of Artificial Intelligence and Robotics, Xi'an Jiaotong University,
    \textsuperscript{\rm 4}The Hong Kong University of Science and Technology (Guangzhou)
}

\begin{document}

\maketitle

\begingroup
\renewcommand{\thefootnote}{\fnsymbol{footnote}}
\footnotetext[1]{Equal contribution.}
\footnotetext[2]{Project Lead.}
\footnotetext[3]{Corresponding Author.}
\endgroup

\begin{abstract}
Automated e-commerce poster design requires both high-quality poster generation and flexible editing of existing designs. However, most existing methods either target end-to-end poster generation or follow multi-stage design pipelines, with limited capability for flexible and precise editing of existing posters. To enable unified generation and editing of e-commerce posters, we introduce Text Patch Generation and Editing, a unified task formulation that treats text patches as atomic units and covers four operations: poster generation, patch addition, patch deletion, and patch modification, with optional reference-guided style control. Based on this, we propose PosterText, a unified model trained with a four-stage curriculum, including text rendering pretraining, instruction-following training, reinforcement learning for preference alignment, and spatial guidance self-distillation for execution refinement. We further construct a large-scale dataset with patch-level annotations and a comprehensive benchmark for evaluation. Extensive experiments demonstrate that PosterText achieves competitive performance against existing generation and editing approaches, validating the effectiveness of the proposed framework.
\end{abstract}

\section{Introduction}

\begin{figure*}[!t]
    \centering
    \includegraphics[width=0.95\textwidth]{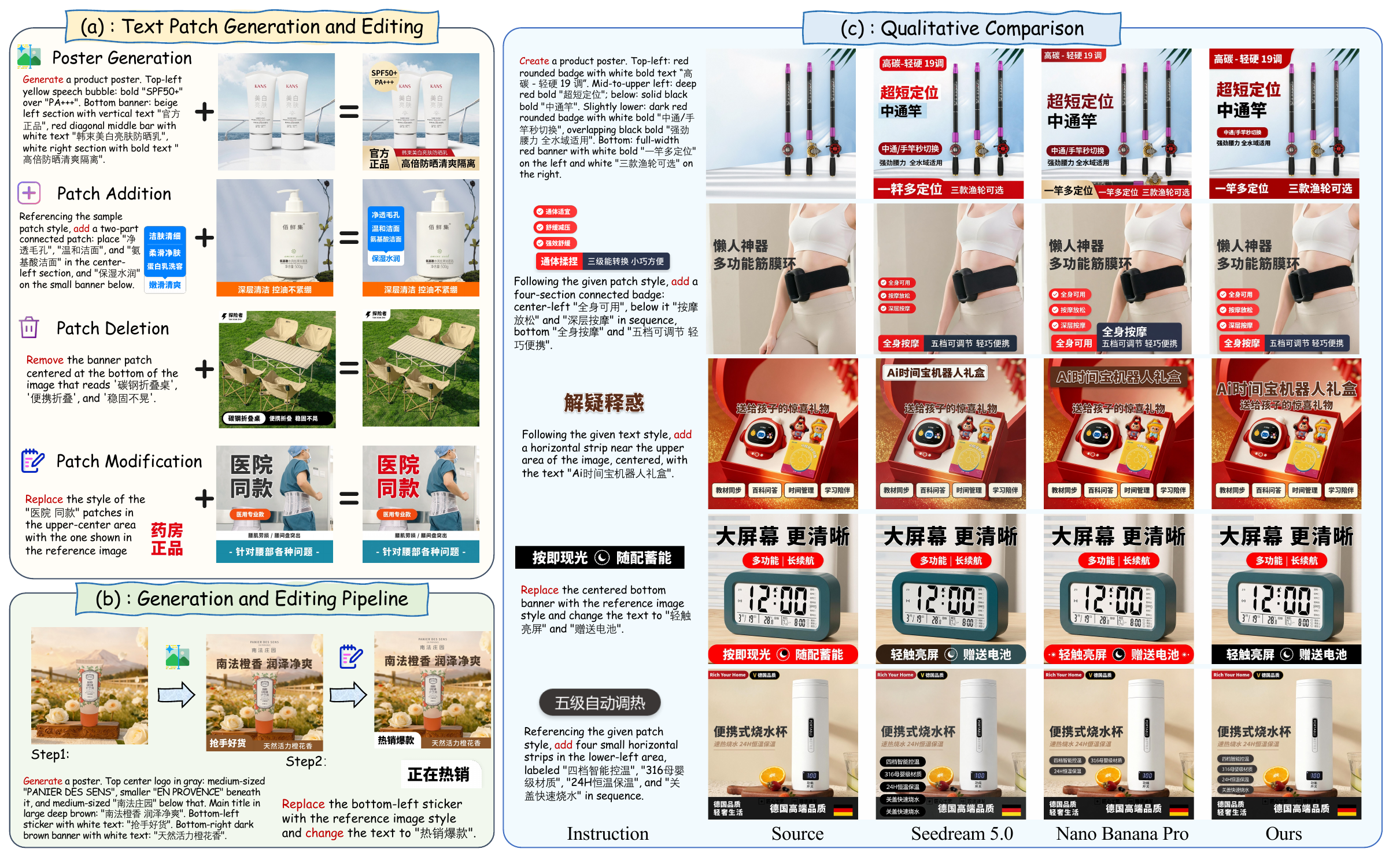}
    \caption{Overview of PosterText framework and qualitative comparison. (a) Four unified operations: poster generation, patch addition, patch deletion, and patch modification. (b) PosterText enables both poster generation and localized editing within a single framework. (c) Qualitative comparison with closed-source baselines.}
    \label{fig:teaser}
\end{figure*}

E-commerce posters are a prevalent form of visual content in online commerce, integrating product images, visual backgrounds, and text elements such as slogans, prices, and brand names~\cite{lin2023autoposter, guo2025repainter, fan2026autopp, qin2026innoads}. These text elements are structured visual units with specific typography, colors, decorations, and spatial layouts, rather than simple strings. In practical scenarios, designers often modify existing posters instead of creating them from scratch, including replacing promotional texts, adjusting typography, adding new campaign elements, or removing outdated information. Therefore, an effective poster automation system should support both high-quality generation and flexible, style-controllable editing based on existing designs.

Existing poster automation methods mainly follow two paradigms. The first adopts a plan-and-render pipeline, which decomposes poster creation into multiple stages such as layout planning and rendering, providing explicit control over the generation process~\cite{li2023planning, hsu2023posterlayout}. However, these methods are constrained by the dependency between stages, where errors can accumulate throughout the pipeline, and local modifications often require re-executing multiple components. The second explores end-to-end diffusion-based generation~\cite{gao2025postermaker, chen2025postercraft, qin2026innoads}, which directly synthesizes complete posters with improved visual quality. Despite their strong generation capability, these methods mainly focus on zero-to-one creation and provide limited support for localized editing of existing designs. As a result, existing approaches remain insufficient for practical poster design workflows, where designers frequently need to iteratively refine existing layouts and introduce new content while preserving visual consistency. Therefore, developing a unified model that supports both poster generation and editing remains challenging. Such a model needs to achieve accurate character-level text rendering, reconcile the different requirements of open-ended generation and precise local editing, and enable reference-guided style control while maintaining overall visual coherence.

To address these challenges, we formulate \textit{Text Patch Generation and Editing}, where text patches serve as the basic units for both generation and editing operations. The proposed framework unifies four operations within a single model: poster generation, patch addition, patch deletion, and patch modification, with optional reference-patch conditioning for style-controlled editing and generation. Based on this formulation, we propose \textbf{PosterText}, a unified model trained with a progressive four-stage curriculum. The model first learns accurate text rendering and patch synthesis, then acquires instruction-driven generation and editing capabilities. Subsequently, reinforcement learning aligns the model with human preferences through text accuracy, aesthetic quality, and background preservation rewards, and Spatial Guidance Self-Distillation (SGSD) further improves execution reliability by transferring mask-guided spatial knowledge to mask-free inference. To facilitate research in this direction, we construct \textbf{PosterText-320K}, a large-scale poster dataset with patch-level structural annotations and automatically generated editing pairs, and further build \textbf{PosterText-Bench} with a comprehensive evaluation protocol.

Our contributions are summarized as follows:
(1) We introduce the task of \textbf{text patch generation and editing}, which treats text patches as atomic units and unifies generation, addition, deletion, and modification with reference-guided style control. Extensive experiments demonstrate the effectiveness of our approach.
(2) We propose \textbf{PosterText}, a unified model trained with a four-stage curriculum covering text rendering, instruction following, reinforcement learning alignment, and self-distillation refinement, enabling both generation and editing within a single framework.
(3) We construct a large-scale dataset and benchmark with an automated data construction pipeline and comprehensive evaluation protocols.

\section{Related Work}

\subsection{E-commerce and Poster Generation}

E-commerce posters are a crucial form of visual content in online commerce, requiring the harmonious integration of products, backgrounds, and marketing texts. Automated poster generation has thus attracted increasing attention, with the key challenge being the creation of visually appealing posters while maintaining product fidelity and text accuracy~\cite{hsu2023posterlayout, chen2025posta, liu2026posterverse, chen2026posteromni}. Early approaches typically follow a plan-and-render paradigm, decomposing generation into layout planning and background rendering stages~\cite{li2023planning}. While these methods provide explicit layout control, the errors introduced across stages and limited text rendering capability hinder their practical applications. Recent works have explored end-to-end generation by jointly modeling text, layout, and visual appearance, achieving impressive performance in generating complete posters from scratch~\cite{gao2025postermaker, chen2025postercraft, qin2026innoads}. However, real-world e-commerce scenarios often require localized modifications to existing designs, such as replacing promotional texts, adjusting typography styles, or adding new campaign elements. Existing approaches primarily focus on zero-to-one generation and lack the ability to perform localized and controllable modifications on specific poster components, making fine-grained editing challenging and limiting their practical usability. Developing a unified model that can both generate high-quality posters and enable flexible, controllable, and fine-grained content editing remains an open challenge.

\begin{figure*}[!t]
    \centering
    \includegraphics[width=0.95\textwidth]{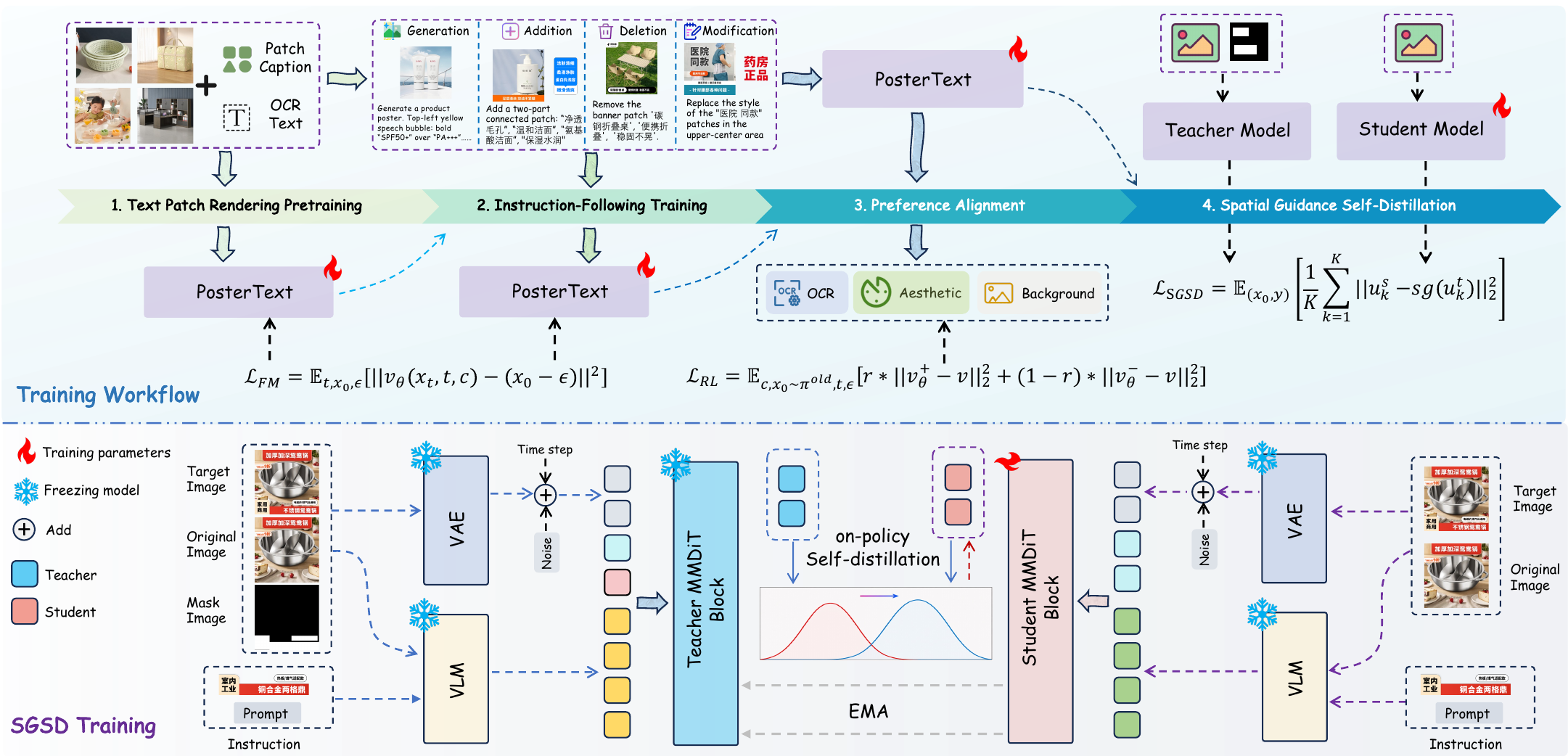}
    \caption{Overview of PosterText. PosterText is trained through four stages: text patch rendering pretraining, instruction-following training, reinforcement learning with human preferences, and Spatial Guidance Self-Distillation~(SGSD). The framework supports four operations: generation, addition, deletion, and modification. For SGSD training, the teacher model uses mask-guided inputs for spatial supervision, while the student learns mask-free localization through on-policy self-distillation.}
    \label{fig:framework}
\end{figure*}

\subsection{Visual Text Rendering and Editing}

Text rendering in images has attracted increasing attention, as diffusion models still struggle with accurate text generation, particularly for complex glyphs~\cite{ma2023glyphdraw, tuo2024anytext2, ma2024chargen, ma2025glyphdraw2, chen2026styletextgen, lu2026easytext}. This challenge is especially critical for e-commerce posters, where text conveys essential product information and marketing semantics. Existing studies have improved text rendering from different perspectives. AnyText~\cite{tuo2024anytext} and FLUX-Text~\cite{lan2025flux} enhanced rendering accuracy through glyph-aware conditioning and improved text-image alignment. Recent works further investigate controllable text generation, with FonTS~\cite{shi2025fonts} and Calligrapher~\cite{ma2025calligrapher} enabling typography and style customization, while UM-Text~\cite{ma2026text} introduced vision-language guidance for instruction-driven text editing. Despite these advances, existing methods mainly treat text as an independent rendering target, overlooking its interaction with the overall poster composition. In practical e-commerce designs, text is tightly coupled with typography, color, decoration, and spatial layout, requiring coordinated generation and editing with surrounding visual elements. To address this limitation, we propose a unified framework that integrates poster generation with patch-level text editing. By incorporating reference patch conditions, our framework enables style-controllable text creation and modification while maintaining visual consistency.

\section{Method}
We present PosterText, a unified framework for text patch generation and editing in e-commerce posters, as shown in Figure~\ref{fig:framework}. Given a product image and a natural-language instruction, PosterText generates or edits posters by treating text patches as atomic units. The model is trained with a multi-stage curriculum that progressively learns text rendering and patch synthesis, extends to multi-task generation and editing, and further improves alignment and execution accuracy through reinforcement learning and self-distillation.

\subsection{Task Formulation}

\noindent
\textbf{Text Patch Definition.} 
We represent a text patch as $p = (t, s, b, r)$, where $t$ denotes the text content, $s$ encodes style attributes (e.g., font, size, and color), $b$ represents an optional decorative background (set to empty for plain text), and $r = (x, y, w, h)$ specifies its spatial region on the canvas. An e-commerce poster $I$ is composed of a product image $I_{\text{prod}}$ and a set of text patches $\mathcal{P} = \{p_1, \ldots, p_n\}$.

\noindent
\textbf{Unified Framework.} 
We model all tasks using a single function $f_\theta$, conditioned on an input image $I_{\text{in}}$, a natural-language instruction $c$, and an optional set of reference patches $\mathcal{R} = \{p_{\text{ref}}^{(1)}, \ldots, p_{\text{ref}}^{(K)}\}$, where $\mathcal{R}=\emptyset$ when no reference patch is provided:

\begin{equation}
I_{\text{out}} = f_\theta(I_{\text{in}}, c, \mathcal{R})
\end{equation}
Specifically, task types are determined by the natural-language instruction $c$: 
(1) \textbf{Poster generation} synthesizes a complete poster from $I_{\text{prod}}$, guided by $c$ describing the desired content and style. 
(2) \textbf{Patch addition} inserts new text patches into an existing poster, where $c$ specifies content and optionally location; style can be further guided by $\mathcal{R}$. 
(3) \textbf{Patch deletion} removes regions specified in $c$. 
(4) \textbf{Patch modification} updates text content or style within a target region defined by $c$, optionally conditioned on $\mathcal{R}$.

\subsection{Stage I: Text Patch Rendering Pretraining}

Accurate text rendering is fundamental for poster generation, as diffusion models often suffer from character-level errors. Directly training instruction following tends to prioritize global layout while degrading text fidelity. We therefore first establish text rendering and patch generation through pretraining.
Stage~I includes two progressive sub-tasks. 
\textbf{(A) Text Rendering:} the model receives prompts containing OCR text and spatial positions, and renders the corresponding glyph sequence. 
\textbf{(B) Patch Reconstruction:} the model reconstructs complete text patches from natural-language descriptions, learning to compose content, style, and layout.

Training starts with Task~A and gradually increases the proportion of Task~B. Both tasks are optimized with the flow matching objective:
\begin{equation}
\mathcal{L}_{\text{FM}} = \mathbb{E}_{t, x_0, \epsilon} \left[ \| v_\theta(x_t, t, c) - (x_0 - \epsilon) \|^2 \right]
\end{equation}

\noindent
where $x_t = (1-t)\epsilon + t \cdot x_0$ is the interpolated sample at time $t$. This stage establishes reliable character-level rendering and then transitions to structured patch synthesis, forming the foundation for subsequent editing tasks.

\subsection{Stage II: Instruction-Following Training}

Building upon the rendering capabilities from Stage~I, the model is further trained to follow instructions and perform diverse operations, including poster generation, patch addition, modification, and deletion. We formulate all tasks with a unified input format $(I_{\text{in}}, c, \mathcal{R})$, where the instruction $c$ specifies the desired operation, target content, and optional style conditions. Training data is constructed by decomposing existing posters into operation tuples, allowing the model to learn different generation and editing behaviors under the same instruction framework. During Stage~II, we jointly train the model with both mask-guided and mask-free samples, allowing it to acquire stronger spatial editing capabilities with explicit masks while supporting mask-free inference.

All operations are jointly optimized with the flow matching objective, with balanced sampling across tasks to ensure stable multi-task learning. After this stage, the model can interpret instructions and perform different poster generation and editing operations, including style-controllable generation and modification with optional reference patches.

\subsection{Stage III: Preference Alignment}

The first two stages optimize pixel-level reconstruction objectives, which are insufficient to capture higher-level perceptual criteria in e-commerce posters, including text readability, visual aesthetics, and background consistency. Building upon the instruction-following capability acquired in Stage~II, we further align the model with human preferences through reinforcement learning.

\noindent
\textbf{Optimization Objective.}
We adopt DiffusionNFT~\cite{zheng2025diffusionnft}, an online RL algorithm compatible with the flow matching framework. Given online rollouts with normalized rewards $r$, the model is optimized by jointly improving high-reward behaviors and suppressing low-reward behaviors through implicit positive and negative policies:

\begin{equation}
\footnotesize
\mathcal{L}_{\text{RL}} = \mathbb{E}_{c, x_0 \sim \pi^{\text{old}}, t, \epsilon}
\left[
r\cdot \|v_\theta^+-v\|_2^2
+
(1-r)\|v_\theta^--v\|_2^2
\right]
\end{equation}

\noindent
where $v=x_0-\epsilon$ denotes the target velocity and $v^{\text{old}}$ is the EMA reference policy. The positive and negative branches are derived from the same velocity field, allowing the model to improve toward preferred outputs while preserving the flow matching formulation.

\noindent
\textbf{Reward Design.}
We design a composite reward tailored to e-commerce poster generation and editing:

\begin{equation}
R=
\alpha R_{\text{text}}
+\beta R_{\text{aesthetic}}
+\delta R_{\text{background}}
\end{equation}

\noindent
where $R_{\text{text}}$ measures OCR-based text accuracy to evaluate character-level rendering fidelity; $R_{\text{aesthetic}}$ assesses the overall visual quality; and $R_{\text{background}}$ measures the preservation of non-edited regions.

\subsection{Stage IV: Spatial Guidance Self-Distillation}

After preference alignment, the model shows significant improvement in overall quality, but still suffers from execution errors in complex scenarios such as ghost text, missing characters, and incorrect operations. These failures mainly arise from insufficient localization and decomposition of complex instructions, making it difficult for the model to accurately determine where and how to perform each operation~\cite{baron2026analysis}. The sparse reward signal in RL only evaluates the final output and cannot effectively attribute global feedback to specific intermediate-step errors.
To improve execution accuracy, we introduce SGSD, which transfers the spatial guidance from mask-guided editing to a mask-free setting. Since explicit masks can provide precise region information and improve editing quality~\cite{guo2026gmo}, SGSD aims to enable the model to achieve comparable execution quality without additional mask inputs.

During training, we construct teacher-student pairs with asymmetric conditions. The teacher receives an additional mask-guided condition $M_t=(I_{\text{in}}, c, \mathcal{R}, \text{mask})$, where the mask provides explicit spatial information for the target operation, while the student receives a mask-free condition $M_s=(I_{\text{in}}, c, \mathcal{R})$. The student samples intermediate states along its own generation trajectory, and the teacher provides velocity predictions at the same states as distillation targets. The training objective is:

\begin{equation}
\mathcal{L}_{\text{SGSD}} = \mathbb{E}_{(x_0, y)} \left[ \frac{1}{K} \sum_{k=1}^{K} \left\| u_k^s - \text{sg}(u_k^t) \right\|_2^2 \right]
\end{equation}
where:
\begin{equation}
u_k^s = \nu_\theta(x_{t_k}^s, t_k, M_s), \quad
u_k^t = \nu_{\bar{\theta}}(x_{t_k}^s, t_k, M_t)
\end{equation}
$\text{sg}(\cdot)$ represents the stop-gradient operation, $K$ denotes the number of intermediate timesteps sampled along each student trajectory, and $\bar{\theta}$ denotes the teacher parameters updated by EMA. Unlike teacher-forcing based SFT~\cite{fu2025dynamic,tan2026meta}, SGSD provides supervision along the student's own generation trajectory, reducing the train-inference discrepancy. The dense intermediate supervision across multiple time steps provides fine-grained correction for intermediate steps that sparse RL rewards cannot reach, further improving execution precision in complex mask-free scenarios.

\begin{figure*}[!t]
    \centering
    \includegraphics[width=0.96\textwidth]{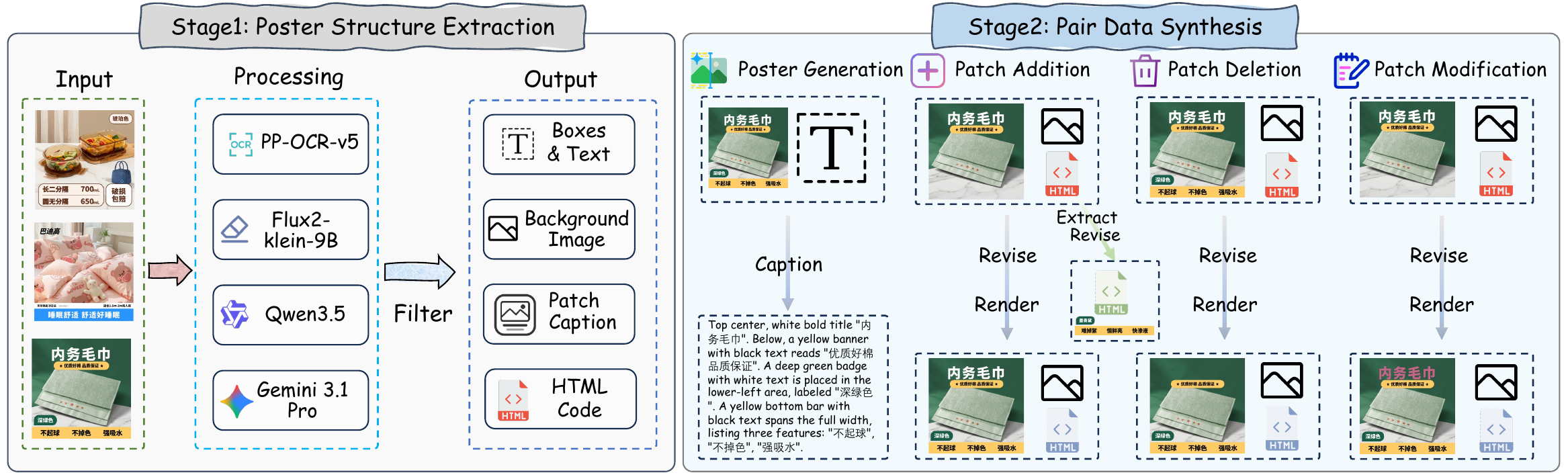}
    \vspace{-1mm}
    \caption{Overview of the data construction pipeline. The pipeline first extracts patch-level structural annotations from raw posters, and then constructs operation pairs for generation, addition, deletion, and modification.}
    \vspace{-1mm}
    \label{fig:data_pipeline}
\end{figure*}

\section{Dataset and Benchmark}

Existing text generation datasets mainly focus on text rendering from scratch, while lacking editing operation pairs and patch-level structural annotations. To support systematic research on text patch generation and editing, we construct \textbf{PosterText-320K}, a large-scale annotated poster dataset containing 120K posters for editing tasks and 200K posters for generation tasks. We further build \textbf{PosterText-Bench} with a comprehensive evaluation protocol. The overall data construction pipeline consists of two stages: first, extracting patch-level structural annotations from raw posters; second, constructing diverse operation pairs.

\subsection{Data Construction Pipeline}

\textbf{Phase I: Poster Structure Extraction.}
We collect a large number of e-commerce posters from public platforms and automatically recover patch-level structural information through a multi-stage pipeline, as illustrated in Figure~\ref{fig:data_pipeline}: (1) text regions are detected and OCR is applied to obtain text content and spatial locations; (2) Qwen3.5-27B~\cite{team2026qwen35} VLM is used to describe visual attributes of each text patch, including font, color, background blocks, and text effects; (3) Flux.2-Klein-9B~\cite{blackforestlabs2025flux2klein} is applied to remove all text patches from posters, producing corresponding product images; (4) Gemini-3.1 Pro~\cite{gemini2026pro3} is used to extract structured HTML representations from the posters; (5) low-quality samples are filtered based on OCR confidence, image resolution, perceptual similarity, and residual text artifacts in the generated product images.

\noindent
\textbf{Phase II: Paired Data Synthesis.}
Based on the extracted poster structures, we construct training pairs for four operations: generation, addition, deletion, and modification. We introduce an HTML-based structured representation that encodes each text patch with its content, visual attributes, and spatial layout, enabling precise patch-level manipulation and reference-based editing. Using the structured representation, we construct training pairs for four operations:
(1) \textbf{Generation:} patch annotations are converted into natural-language descriptions to create product-conditioned poster generation pairs.
(2) \textbf{Addition:} existing patches are removed and re-rendered to construct addition pairs, with HTML attributes providing reference patch information for style-conditioned generation.
(3) \textbf{Deletion:} target patches are removed according to their structural locations to generate deletion pairs.
(4) \textbf{Modification:} patch content or visual attributes are changed while preserving the original layout, with HTML attributes enabling reference-guided style-controlled editing.

\subsection{Evaluation Protocol}

We evaluate model outputs from three perspectives: text fidelity, background preservation, and editing quality.

\noindent
\textbf{Text Fidelity.}
Sentence Accuracy (Sen. Acc) and Normalized Edit Distance (NED) are used to evaluate text rendering accuracy, measuring sentence-level exact correctness and character-level similarity between generated and ground-truth texts, respectively.

\noindent
\textbf{Background Preservation.}
Masked-LPIPS is used to evaluate background preservation by measuring perceptual differences between input and output images in non-edited regions. For poster generation, a VLM-based scorer is instead used to assess whether the product subject and content are faithfully preserved in the generated output.

\noindent
\textbf{Editing Quality.}
A vision-language model is used to evaluate editing quality from two aspects: instruction following, which measures whether the requested operation is correctly executed, and visual quality, which evaluates the coherence, style consistency, and naturalness of the edited results.

\begin{figure*}[!t]
    \centering
    \includegraphics[width=0.95\textwidth]{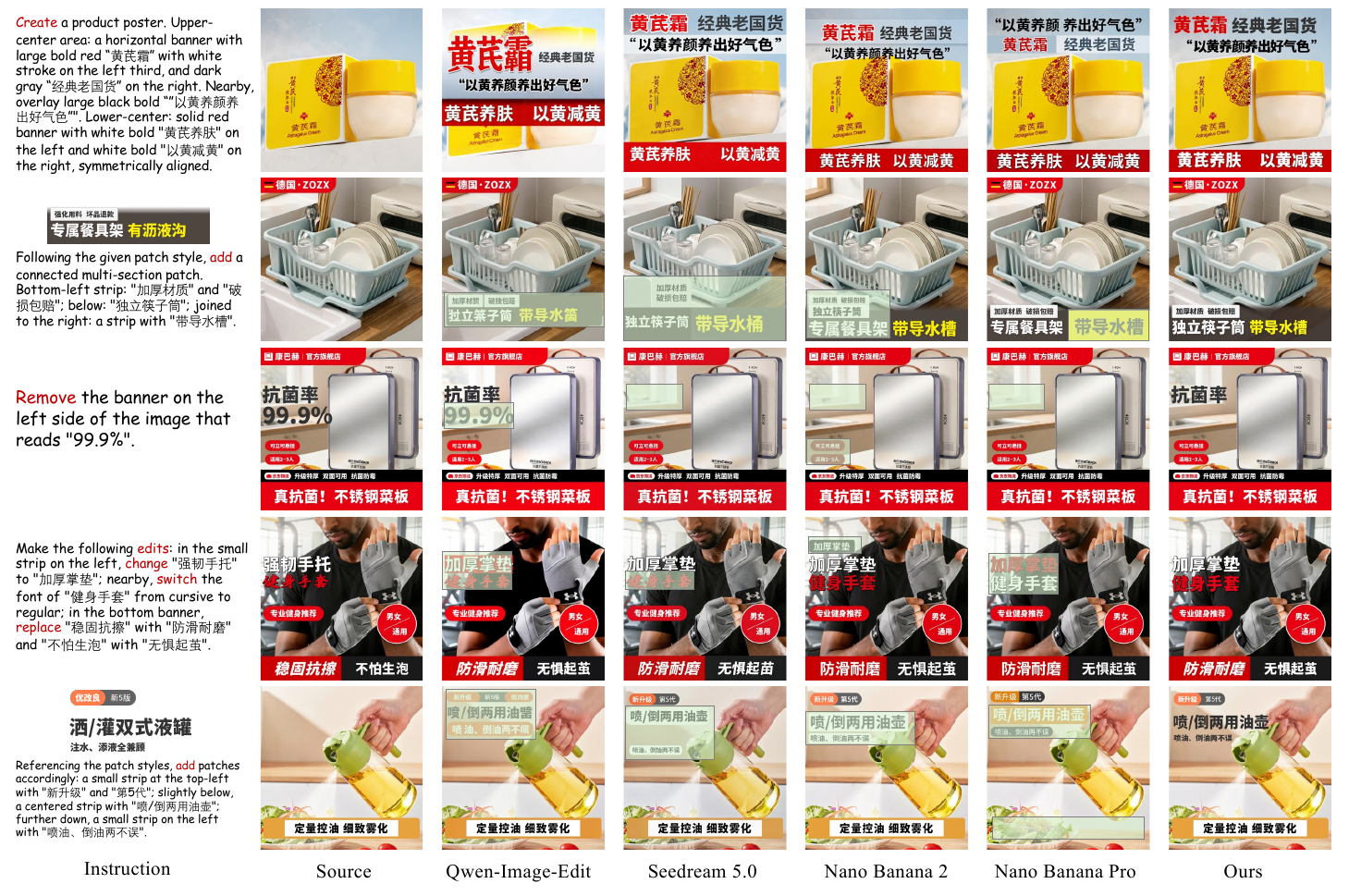}
    \vspace{-2mm}
    \caption{Visual comparison of PosterText with existing methods.}
    \vspace{-1mm}
    \label{fig:visual_comparison}
\end{figure*}

\begin{table*}[t]
\centering
\resizebox{\textwidth}{!}{%
\begin{tabular}{l ccccc ccc ccccc ccc}
\toprule
& \multicolumn{5}{c}{\textbf{Patch Addition}}
& \multicolumn{3}{c}{\textbf{Patch Deletion}}
& \multicolumn{5}{c}{\textbf{Patch Modification}}
& \multicolumn{3}{c}{\textbf{Poster Generation}} \\
\cmidrule(lr){2-6}\cmidrule(lr){7-9}\cmidrule(lr){10-14}\cmidrule(lr){15-17}
\textbf{Method}
  & \textbf{S.A.} & \textbf{NED} & \textbf{M-L.} & \textbf{I.F.} & \textbf{V.Q.}
  & \textbf{M-L.} & \textbf{I.F.} & \textbf{V.Q.}
  & \textbf{S.A.} & \textbf{NED} & \textbf{M-L.} & \textbf{I.F.} & \textbf{V.Q.}
  & \textbf{I.F.} & \textbf{V.Q.} & \textbf{B.P.} \\
\midrule
\multicolumn{17}{l}{\textit{Closed-Source Methods}} \\
\midrule
Seedream 5.0
  & 0.403 & 0.585 & 0.195 & 4.40 & 4.64
  & 0.122 & \underline{4.68} & 4.69
  & 0.735 & 0.884 & 0.102 & 4.33 & \underline{4.65}
  & 4.10 & 4.50 & 4.80 \\
Nano Banana 2
  & 0.500 & 0.635 & \underline{0.064} & 4.46 & \underline{4.67}
  & \underline{0.051} & 4.62 & \underline{4.84}
  & \textbf{0.819} & \textbf{0.907} & \underline{0.034} & \underline{4.47} & \textbf{4.80}
  & \textbf{4.63} & \underline{4.76} & 4.86 \\
Nano Banana Pro
  & \underline{0.543} & \underline{0.679} & 0.184 & \textbf{4.55} & \textbf{4.75}
  & 0.061 & 4.66 & \textbf{4.91}
  & \underline{0.815} & 0.891 & 0.102 & 4.36 & 4.64
  & \underline{4.58} & \textbf{4.88} & \textbf{4.92} \\
\midrule
\multicolumn{17}{l}{\textit{Open-Source Methods}} \\
\midrule
FireRed-Image-Edit
  & 0.309 & 0.481 & 0.497 & 3.53 & 3.61
  & 0.185 & 4.29 & 4.04
  & 0.673 & 0.803 & 0.305 & 3.71 & 3.78
  & 3.86 & 4.21 & 4.42 \\
Flux.2-Klein-9B
  & 0.163 & 0.284 & 0.100 & 1.51 & 2.55
  & 0.098 & 4.02 & 4.57
  & 0.375 & 0.573 & 0.062 & 2.16 & 3.02
  & 3.12 & 3.68 & 4.24 \\
Qwen-Image-Edit
  & 0.378 & 0.563 & 0.647 & 3.31 & 3.31
  & 0.323 & 3.08 & 2.76
  & 0.602 & 0.769 & 0.280 & 3.12 & 3.36
  & 3.58 & 4.06 & 2.92 \\
\midrule
PosterText
  & \textbf{0.763} & \textbf{0.899} & \textbf{0.039} & \underline{4.51} & 4.61
  & \textbf{0.040} & \textbf{4.70} & 4.78
  & 0.771 & \underline{0.902} & \textbf{0.028} & \textbf{4.53} & 4.64
  & 4.21 & 4.46 & \underline{4.88} \\
\bottomrule
\end{tabular}%
}
\caption{Quantitative results on PosterText-Bench. \textbf{Bold} and \underline{underline} indicate the best and second-best results, respectively. S.A.: Sentence Accuracy; M-L.: Masked-LPIPS; I.F.: Instruction Following; V.Q.: Visual Quality; B.P.: Background Preservation. Higher~$\uparrow$ is better for S.A., NED, I.F., V.Q., and B.P., while lower~$\downarrow$ is better for M-L.}

\label{tab:main_comparison}
\end{table*}

\section{Experiments}

\noindent
\textbf{Implementation Details and Baselines.}
We build PosterText upon Qwen-Image-Edit and fine-tune it with LoRA. Stage~I is trained for 50K steps with a cosine learning rate schedule and an initial learning rate of $1\times10^{-4}$. The sampling ratio between text rendering task and patch reconstruction task gradually changes from 4:1 to 1:1 during the first 20K steps, with LoRA rank set to 128. Stage~II is trained for 10K steps for instruction following. Stage~III applies DiffusionNFT for RL alignment with 200 training steps, while Stage~IV performs self-distillation with $K=20$ and 500 training steps. More implementation details are provided in the supplementary material. 

For evaluation, we compare PosterText with representative open-source and closed-source methods. The open-source baselines include Flux.2-Klein-9B~\cite{blackforestlabs2025flux2klein}, Qwen-Image-Edit~\cite{wu2025qwenimage}, and FireRed-Image-Edit~\cite{zhou2025fireedit}. The closed-source baselines include Seedream~5.0~\cite{seedream2025seedream}, Nano Banana 2~\cite{gemini2026nano2}, and Nano Banana Pro~\cite{gemini2026nanopro}.

\subsection{Experimental Results}

We conduct a unified evaluation of the above methods across all tasks on PosterText-Bench, with comprehensive results summarized in Table~\ref{tab:main_comparison}.
Overall, PosterText achieves strong performance in text fidelity, background preservation, and instruction following, substantially outperforming existing open-source baselines and even surpassing closed-source models on key metrics. 
In terms of visual quality, PosterText narrows the gap with closed-source systems, demonstrating competitive generation and editing quality.
As shown in Figure~\ref{fig:teaser}~(c) and~\ref{fig:visual_comparison}, PosterText can better leverage complex style references to generate corresponding text patches while maintaining accurate text rendering and clean backgrounds. In contrast, open-source baselines often produce blurry or corrupted characters, while closed-source models generally achieve higher visual quality but may modify non-target regions, introduce unintended content, or fail to preserve the original patch semantics faithfully.

For Patch Addition, PosterText achieves the best text fidelity among all methods, with Sen.Acc of 0.763 and NED of 0.899, outperforming the closed-source systems. It also obtains the lowest Masked-LPIPS score, demonstrating that PosterText can insert new patches while effectively preserving the original poster structure.
For Patch Deletion, which mainly evaluates region localization and background restoration rather than text rendering, PosterText achieves competitive Masked-LPIPS and instruction-following scores, indicating more accurate spatial control and cleaner content removal.
Patch Modification represents the most challenging setting, as it requires simultaneous text regeneration, precise target localization, and preservation of surrounding regions. PosterText achieves strong performance in background preservation and instruction following, demonstrating effective localized editing while maintaining overall visual consistency. 
Some closed-source models achieve higher text fidelity on this task, which may benefit from large-scale proprietary training data and broader text rendering supervision.

In the Poster Generation task, closed-source models maintain advantage in instruction following and visual quality, benefiting from larger-scale training data, stronger aesthetic alignment, and more capable foundation models. Meanwhile, PosterText achieves comparable performance in product fidelity, demonstrating that the region control and content preservation abilities learned from editing tasks can effectively transfer to the generation setting. Despite their strong performance, closed-source models still face practical limitations in e-commerce deployment, including high costs for large-scale generation due to pay-per-use pricing, limited adaptability caused by the lack of fine-tuning access, and potential concerns regarding commercial data privacy when uploading product images to external services. In contrast, PosterText provides a more controllable and deployment-friendly solution for practical poster generation scenarios.

The comparison results demonstrate that PosterText addresses the key challenge of this work: unifying poster generation and fine-grained patch-level editing within a single model. Across three editing tasks, PosterText achieves strong performance in text fidelity and region preservation, two essential requirements often underemphasized by existing generation-oriented methods. By combining accurate text rendering with precise local control under a unified instruction framework, PosterText bridges the gap between text rendering models and general-purpose image editing systems. For poster generation, the performance gap with closed-source models mainly reflects differences in training scale and optimization focus, while the strong product fidelity of PosterText confirms that editing-oriented region control can effectively transfer to generation scenarios.

\subsection{Ablation Study}

\begin{table}[t]
\centering
\small
\setlength{\tabcolsep}{6pt}
\begin{tabular}{lrrrrr}
\toprule
\textbf{Model} & \textbf{S.A.}$\uparrow$ & \textbf{NED}$\uparrow$ & \textbf{M-L.}$\downarrow$ & \textbf{I.F.}$\uparrow$ & \textbf{V.Q.}$\uparrow$ \\
\midrule
Base Model     & 0.490 & 0.666 & 0.417 & 3.27 & 3.37 \\
\midrule
\multicolumn{6}{l}{\textit{+SFT Stage}} \\
Stage II-only         & 0.679 & 0.786 & 0.088 & 3.63 & 3.81 \\
Stage I(no cur.)+II   & 0.702 & 0.831 & 0.061 & 4.00 & 4.12 \\
Stage I + II          & 0.712 & 0.866 & 0.053 & 4.03 & 4.24 \\
\midrule
+SFT+RL            & 0.735 & 0.883 & 0.037 & 4.22 & 4.51 \\
\midrule
+SFT+RL+SGSD          & 0.767 & 0.901 & 0.036 & 4.49 & 4.62 \\
\bottomrule
\end{tabular}%

\caption{Ablation study of different training stages, averaged across all tasks.}
\label{tab:ablation}
\end{table}

To evaluate the contribution of each training stage, we conduct a systematic ablation study using task-averaged metrics, with results summarized in Table~\ref{tab:ablation}. Specifically, Sen.ACC and NED are averaged over the Addition and Modification tasks; Masked-LPIPS is averaged over Addition, Deletion, and Modification; instruction following and visual quality scores are averaged across all four tasks.

\noindent\textbf{Analysis of the SFT stage.}
Without task-specific training, the base model achieves only 0.490 S.A. and 0.417 M-L., indicating limited capability for patch-level editing. Stage~II-only substantially improves performance, increasing S.A. to 0.679 and reducing M-L. to 0.088, demonstrating that instruction-following training establishes basic editing capabilities. 
Adding Stage~I pretraining (Stage~I(no~cur.)+II) further improves NED from 0.786 to 0.831, confirming that text rendering pretraining strengthens character-level representation. Introducing the curriculum sampling strategy (Stage~I+II) brings additional but modest improvements, mainly by providing a more stable initialization for subsequent RL optimization.

\noindent\textbf{Analysis of the RL stage.}
Building on full SFT, RL further improves editing performance and visual quality. This demonstrates that pixel-level reconstruction objectives alone cannot fully capture human preferences, such as text correctness, aesthetic quality, and instruction compliance. The multi-dimensional reward design effectively guides different aspects of optimization: OCR-based rewards improve text fidelity, aesthetic rewards enhance visual quality, and instruction-based rewards improve task execution. The significant reduction in M-L. (30.2\%) further indicates that the background preservation reward explicitly constrains editing boundaries, reducing unintended modifications.

\noindent\textbf{Analysis of the SGSD stage.}
SGSD further enhances text accuracy and instruction following through dense trajectory-level supervision, which corrects intermediate-step errors beyond the reach of sparse reward signals. The improvement in instruction following indicates that spatial knowledge from the mask-guided teacher can be effectively transferred to mask-free inference for more accurate region localization. Since RL already provides strong background preservation, SGSD mainly improves text rendering precision and execution reliability. Together, RL and SGSD offer complementary benefits: RL optimizes overall editing quality, while SGSD improves execution accuracy.

\section{Conclusion}
\vspace{2px}
In this work, we have introduced the task of Text Patch Generation and Editing, which treats text patches as fundamental units for automated e-commerce poster design and unifies poster generation, patch addition, deletion, and modification within a single framework. Based on this formulation, we propose PosterText, a unified model trained with a four-stage curriculum that sequentially develops text rendering, instruction following, preference alignment, and execution refinement capabilities. Extensive experiments demonstrate that PosterText achieves competitive performance in text fidelity, background preservation, and instruction execution, validating effectiveness of unified generation and editing.

\bibliography{aaai2027}


\end{document}